\pdfoutput=1

\documentclass[11pt]{article}

\usepackage{acl}

\usepackage{times}
\usepackage{latexsym}

\usepackage[T1]{fontenc}

\usepackage[utf8]{inputenc}

\usepackage{microtype}
\usepackage{amsmath}
\usepackage{graphicx}
\usepackage{tabularx}

\newcommand{\src}{\ensuremath{\mathbf{f}}} 
\newcommand{\trg}{\ensuremath{\mathbf{e}}} 

\title{Joint Generation of Captions and Subtitles with Dual Decoding}

\author{Jitao Xu$^{\dag}$\ \  François Buet$^{\dag}$\ \  Josep Crego$^{\ddag}$\ \  Elise Bertin-Lemée$^{\ddag}$\ \  François Yvon$^{\dag}$ \\ \\
$^\dag$Universit\'e Paris-Saclay, CNRS, LISN, 91400, Orsay, France \\
$^\ddag$SYSTRAN, 5 rue Feydeau, 75002 Paris, France \\
\texttt{\{firstname.lastname\}@\{$^{\dag}$limsi.fr,$^{\ddag}$systrangroup.com\}} \\}

\begin{document}
\maketitle

\begin{abstract}
As the amount of audio-visual content increases, the need to develop automatic captioning and subtitling solutions to match the expectations of a growing international audience appears as the only viable way to boost throughput and lower the related post-production costs. Automatic captioning and subtitling often need to be tightly intertwined to achieve an appropriate level of consistency and synchronization with each other and with the video signal. In this work, we assess a dual decoding scheme to achieve a strong coupling between these two tasks and show how adequacy and consistency are increased, with virtually no additional cost in terms of model size and training complexity. 
\end{abstract}

\section{Introduction\label{sec:intro}}

As the amount of online audio-visual content continues to grow, the need for captions and subtitles\footnote{We use `caption' to refer to a text written in the same language as the audio and `subtitle' when translated into another language. Captions, which are often meant for viewers with hearing difficulties, and subtitles, which are produced for viewers with an imperfect command of the source language, may have slightly different traits, that we ignore here.} in multiple languages also steadily increases, as it widens the potential audience of these contents.

\begin{figure}[!ht]
  \center
  \includegraphics[width=0.7\columnwidth]{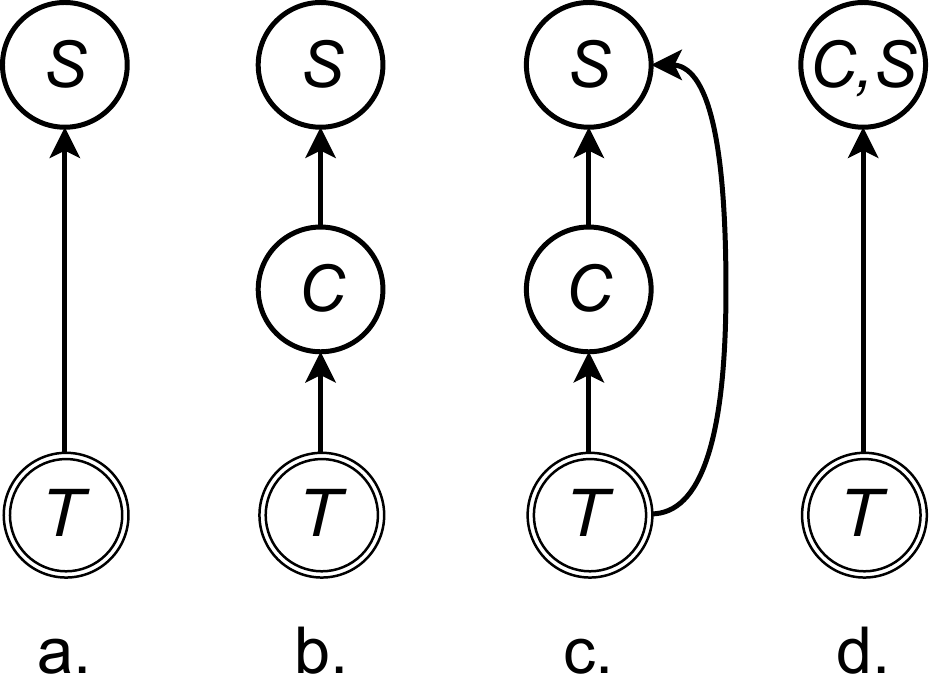}
  \caption{A graphical view of various captioning and subtitling strategies. T refers to transcripts. C and S respectively denote captions and subtitles.\label{fig:strategy}}
\end{figure}

Both activities are closely related: human subtitle translators often generate subtitles directly based on the original captions without viewing or listening to the original audio/video file. This strategy however runs the risk of amplifying, in the subtitle approximations, simplifications or errors present in the captioning.
It may even happen that both texts need to be simultaneously displayed on screen:
for instance, in countries with several official languages, or to help foreign language learners. This means that captions and subtitles need to be consistent not only with the video content, but also with each other. It also implies that they should be synchronized \citep{Karakanta21flexibility}. 
Finally, even in scenarios where only subtitles would be needed, generating captions at the same time may still help to better check the correctness of subtitles.

Early approaches to automatic subtitling (e.g.\ \citealp{Piperidis04multimodal}) also assumed a pipeline architecture (Figure~\ref{fig:strategy}~(b)), where subtitles are translated from captions derived from automatic speech transcripts. A recent alternative (Figure~\ref{fig:strategy} (a)), which mitigates cascading errors, is to independently perform captioning and subtitling in an end-to-end manner \citep{Liu20adapting,Karakanta20answer}; the risk however is to generate inconsistencies (both in alignment and content) between the two textual streams. This approach might also be limited by the lack of appropriate training resources \citep{Sperber20speech}. Various ways to further strengthen the interactions between these tasks by sharing parameters or loss terms are evaluated by \citet{Sperber20consistent}. Figure~\ref{fig:strategy}~(c) illustrates these approaches.

\begin{table*}[ht]
  \center
  \scalebox{0.9}{
  \begin{tabularx}{1.1\textwidth}{l|X}
  \hline
  Transcript & \textbf{i 'm} combining specific types of signals \textbf{the} mimic how our body \textbf{response} to \textbf{in an injury} to help us regenerate \\
  Caption & \textbf{I'm} combining specific types of signals \textbf{[eob]} \textbf{that} mimic how our body \textbf{responds} to \textbf{injury} \textbf{[eol]} to help us regenerate\textbf{.} \textbf{[eob]} \\
  Subtitle & Je combine différents types de signaux [eob] qui imitent la réponse du corps [eol] aux blessures pour nous aider à guérir. [eob] \\
  \hline
  \end{tabularx}
  }
  \caption{Example of a triplet (transcript, caption, subtitle) from our tri-parallel data. Differences between transcript and caption are in bold.}\label{tab:task}
\end{table*}

In this work, we explore an even tighter integration consisting of \emph{simultaneously generating both captions and subtitles} from automatic speech recognition (ASR) transcripts \emph{using one single dual decoding process} \citep{Zhou19synchronous,Wang19synchronously,Le20dual,He21synchronous,Xu21dual}, illustrated in Figure~\ref{fig:strategy}~(d). Generally speaking, automatically turning ASR transcripts into full-fledged captions involves multiple changes, depending on the specification of the captioning task. In our case, this transformation comprises four main aspects: segmentation for display (via tag insertion), removal of certain features from spoken language (eg.\ fillers, repetitions or hesitations), ASR errors correction, and punctuation prediction. The transcript-to-subtitle task involves the same transformations, with an additional translation step to produce text in another language. Table~\ref{tab:task} illustrates the various transformations that occur between input transcripts and the corresponding output segments. 

As our experiments suggest, a tighter integration not only improves the quality and the consistency of captions and subtitles, but it also enables a better use of all available data, \emph{with hardly any impact on model size or training complexity}. 
Our main contributions are the following:
(i) we show that simultaneously generating captions and subtitles can improve performance in both languages, reporting significant improvements in BLEU score with respect to several baselines;
(ii) we initialize dual decoder from a standard encoder-decoder model trained with large scale data, thereby mitigating the data scarcity problem;
(iii) we explore a new parameter sharing scheme, where the two decoders share all their parameters, and achieve comparable performance at a much reduced model size in our experimental conditions; 
(iv) using $2$-round decoding, we show how to alleviate the exposure bias problem observed in dual decoding, leading to a clear boost in performance.

\section{Dual Decoding\label{sec:dual}}

\subsection{Model\label{ssec:model}}

In a nutshell, dual decoding aims to generate two output sentences $\trg^1$ and $\trg^2$ for each input sentence $\src$. This means that instead of having two independent models (Eq.~\eqref{eq:indepgen}), the generation of each target is influenced by the other output (Eq.~\eqref{eq:jointgen}): 
\begin{align}
  P(\trg^1,\trg^2|\src) &= \prod_{t=1}^{T} P(\trg^{1}_{t}|\src, \trg^{1}_{<t}) P(\trg^{2}_{t}|\src, \trg^{2}_{<t}) \label{eq:indepgen} \\
  P(\trg^1,\trg^2|\src) &= \prod_{t=1}^{T} P(\trg^{1}_{t}|\src, \trg^{1}_{<t}, \trg^{2}_{<t}) \times \nonumber \\
                        & \qquad\quad P(\trg^{2}_{t}|\src, \trg^{1}_{<t}, \trg^{2}_{<t}), \label{eq:jointgen}
\end{align}
where $T = max(|\trg^1|, |\trg^2|)$.

In our experiments, ASR transcripts are considered as the source language while captions and subtitles are the two target languages \citep{Wang19synchronously,He21synchronous,Xu21dual}. The dual decoder model has also been proposed in several application scenarios other than multi-target translation such as bi-directional translation \citep{Zhou19synchronous,Zhang20synchronous,He21synchronous}, and also to simultaneously generate transcripts and translations from the audio source \citep{Le20dual}.

To implement the interaction between the two decoders, we mostly follow \citet{Le20dual} and \citet{Xu21dual} who add a decoder cross-attention layer in each decoder block, so that the hidden states of previous layers of each decoder $H_l^1$ and $H_l^2$ can attend to each other.
The decoder cross-attention layers take the form:\footnote{We define the $\operatorname{Attention}(Q,K,V)$ function as in \citep{Vaswani17attention} as a function of three arguments standing respectively for Query, Key and Value.} 
\begin{equation*}
  \begin{split}
  H^1_{l+1} &= \operatorname{Attention}(H^1_l, H^2_l, H^2_l) \\
  H^2_{l+1} &= \operatorname{Attention}(H^2_l, H^1_l, H^1_l) \,
  \end{split}
  \label{eq:mutual}
\end{equation*}
Both decoders are thus fully synchronous since each requires the hidden states of the other to compute its own hidden states.

\subsection{Sharing Decoders\label{ssec:sharing}}

One weakness of the dual decoder model is that it contains two separate decoders, yielding an increased number of parameters ($\times1.6$ in our models w.r.t. standard translation models). Inspired by the idea of tying parameters in embedding matrices \citep{Inan17tying,Press17using}, we extend the dual decoder model by \emph{sharing all the parameters matrices in the two decoders}: in this way, the total number of parameters remains close to that of a standard translation model ($\times1.1$), since the only increase comes from the additional decoder cross-attention layer. 
When implementing inference with this multilingual shared decoder, we prefix each target sentence with a tag indicating the intended output (captioning or subtitling).

\subsection{Training and Fine-tuning\label{ssec:train}}

The dual decoder model is trained using a joint loss combining the log-likelihood of the two targets:
\begin{equation*}
  \begin{split}
  L(\theta) = \sum_{D} (\sum_{t=1}^{|\trg^1|}\log P(\trg_t^1 | \trg^1_{<t}, \trg^2_{<t}, \src; \theta) \\
  + \sum_{t=1}^{|\trg^2|}\log P(\trg^2_t|\trg^2_{<t}, \trg^1_{<t}, \src; \theta)) \,,
  \end{split}
\end{equation*}
where $\theta$ represents the set of parameters.
Training this model requires triplets of instances associating one source with two targets. Such resources are difficult to find and the largest tri-parallel open source corpus we know of is the MuST-Cinema dataset \citep{Karakanta20must}, which is clearly smaller than what exists to separately train automatic transcription or translation systems.

In order to leverage large scale parallel translation data for English-French, we adopt a fine-tuning strategy where we initially pre-train a standard (encoder-decoder) translation model using all available resources, which serves to initialize the parameters of our dual decoder model.
As the dual decoder network employs two decoders with shared parameters, we use also the decoder of the pre-trained model to initialize this subnetwork.
Fine-tuning is performed on a tri-parallel corpus.
We discuss the effect of decoder initialization in Section~\ref{sssec:ft}.
Finally, for all fine-tuned models, the decoder cross-attention layer which binds the two decoders together is always randomly initialized. 

\section{Experiments\label{sec:exp}}

\subsection{Datasets and Resources \label{ssec:data}}

For our experiments, we use MuST-Cinema\footnote{\url{https://ict.fbk.eu/must-cinema/}} \citep{Karakanta20must}, a multilingual Speech-to-Subtitles corpus compiled from TED talks, in which subtitles contain additional segmentation tags indicating changes of screen ([eob]) or line ([eol]).
Our experiments consider the translation from English (EN) into French (FR). 
Our tri-parallel data also includes a pre-existing unpunctuated ASR output generated by \citet{Karakanta20answer}, which achieves a WER score of 39.2\% on the MuST-Cinema test set speech transcripts (details in Appendix~\ref{sec:details}).
For pre-training, we use all available WMT14 EN-FR data. During fine-tuning, we follow the recommendations and procedures of \citet{Zhou19synchronous,Wang19synchronously,He21synchronous,Xu21dual}, and use synthetic tri-parallel data, in which we alternatively replace one of the two target side references by hypotheses generated from the baseline system for the corresponding direction via forward-translation. For more details about synthetic tri-parallel data generation, we refer to \citep{Zhou19synchronous,Xu21dual}.
We tokenize all data with Moses scripts and use a shared source-target vocabulary of 32K Byte Pair Encoding units \citep{Sennrich16BPE} learned with \texttt{subword-nmt}.\footnote{\url{https://github.com/rsennrich/subword-nmt}}

\subsection{Experimental Settings\label{ssec:settings}}

We implement the dual decoder model based on the Transformer \citep{Vaswani17attention} model using \texttt{fairseq}\footnote{\url{https://github.com/pytorch/fairseq}} \citep{Ott19fairseq}.\footnote{Our implementation is open-sourced at \url{https://github.com/jitao-xu/dual-decoding}} All models are trained until no improvement is found for $4$ consecutive checkpoints on the development set, except for the EN$\to$FR pre-trained translation model which is trained during $300$k iterations (further details in Appendix~\ref{sec:expdetails}). 
We mainly measure performance with SacreBLEU \citep{Post18sacrebleu};\footnote{BLEU+case.mixed+numrefs.1+smooth.exp+tok.13a+ version.1.5.1} TER and BERTScores \citep{Zhang20BERTScore} are also reported in Appendix~\ref{sec:metric}.
Segmentation tags in subtitles are taken into account and BLEU scores are computed over full sentences. 
In addition to BLEU score, measuring the consistency between captions and subtitles is also an important aspect.  We reuse the structural and lexical consistency score proposed by \citet{Karakanta21flexibility}. \emph{Structural consistency} measures the percentage of utterances having the same number of blocks in both languages, while \emph{lexical scores} count the proportion of words in the two languages that are aligned in the same block (refer to Appendix~\ref{sec:consistency} for additional details).

We call the dual decoder model \texttt{dual}. Baseline translation models trained separately on each direction ($\operatorname{T}_{\operatorname{en}}$$\to$$\operatorname{C}_{\operatorname{en}}, \operatorname{T}_{\operatorname{en}}$$\to$$\operatorname{S}_{\operatorname{fr}}$) are denoted by \texttt{base}.
To study the effectiveness of dual decoding, we mainly compare \texttt{dual} with a \texttt{pipeline} system.
The latter uses the \texttt{base} model to produce captions which are then translated into subtitles using an independent system trained to translate from caption to subtitle ($\operatorname{T}_{\operatorname{en}}$$\to$$\operatorname{C}_{\operatorname{en}}$$\to$$\operatorname{S}_{\operatorname{fr}}$).

Like the \texttt{dual} model, \texttt{base} and \texttt{pipeline} systems also benefit from pre-training. For the former, we pre-train the direct transcript-to-subtitle translation model ($\operatorname{T}_{\operatorname{en}}$$\to$$\operatorname{S}_{\operatorname{fr}}$); for \texttt{pipeline}, the caption-to-subtitle model ($\operatorname{C}_{\operatorname{en}}$$\to$$\operatorname{S}_{\operatorname{fr}}$) is pre-trained, while the first step ($\operatorname{T}_{\operatorname{en}}$$\to$$\operatorname{C}_{\operatorname{en}}$) remains as in the \texttt{base} system. Note that all fine-tuned systems start with the same model pre-trained using WMT EN-FR data.

\subsection{Main Results\label{ssec:results}}

\begin{table}[ht]
  \center
  \scalebox{0.85}{
  \begin{tabular}{l|ccc|cc}
  \hline
  & \multicolumn{3}{c|}{BLEU} & \multicolumn{2}{c}{Consistency} \\
  Model & EN   &  FR   & Avg  & Struct.\ & Lex.\ \\
  \hline
  \texttt{base} & 55.7 & 23.9 & 39.8 & 55.3 & 70.7 \\
  \texttt{base} +FT & 55.7 & 24.9 & 40.3 & 54.5 & 71.4 \\
  \texttt{pipeline} & 55.7 & 23.6 & 39.7 & 95.7 & 96.0 \\
  \texttt{pipeline} +FT & 55.7 & 24.2 & 40.0 & 98.4 & 98.3 \\
  \texttt{dual} +FT & \textbf{56.9} & 25.6 & \textbf{41.3} & 65.1 & 79.1 \\
  \texttt{share} +FT & 56.5 & \textbf{25.8} & 41.2 & \textbf{66.7} & \textbf{80.0} \\
  \hline
  \end{tabular}
  }
  \caption{BLEU scores for captions (EN) and subtitles (FR), with measures of structural and lexical consistency between the two hypotheses. These scores are in percentage (higher is better). The \texttt{base} and \texttt{pipeline} settings are trained from scratch with original data. \texttt{share} refers to tying all decoder parameters.\label{tab:res}}
\end{table}

We only report in Table~\ref{tab:res} the performance of the two baselines and fine-tuned (+FT) models, as our preliminary experiments showed that training the dual decoder model with only tri-parallel data was not optimal. 
The BLEU score of the \textit{do nothing} baseline, which copies the source ASR transcripts to the output, is 28.0, which suggests that the captioning task actually involves much more transformations than simply inserting segmentation tags.
We see that fine-tuning improves subtitles generated by \texttt{base} and \texttt{pipeline} systems by $\sim$$1$ BLEU. Our \texttt{dual} decoder model, after fine-tuned using synthetic tri-parallel data, respectively outperforms \texttt{base}+FT by $0.7$ BLEU, and \texttt{pipeline}+FT by $1.4$ BLEU. Sharing all parameters of both decoders yields further increase of $0.2$ BLEU, with about one third less parameters. 

We also measure the structural and lexical consistency between captions and subtitles generated by our systems (see Table~\ref{tab:res}).
As expected, \texttt{pipeline} settings always generate very consistent pairs of captions and subtitles, as subtitles are direct translations of the captions; all other methods generate both outputs from the ASR transcripts. \texttt{dual} models do not perform as well, but are still able to generate captions and subtitles with a much higher structural and lexical consistency between the two outputs than in the \texttt{base} systems. \citet{Xu21dual} show that dual decoder models generate translations that are more consistent in content. We further show here that our \texttt{dual} models generates hypotheses which are also more consistent in structure. Examples output captions and subtitles are in Appendix~\ref{sec:example}.

\subsection{Analyses and Discussions\label{ssec:analysis}}

\subsubsection{The Effect of Fine-tuning\label{sssec:ft}}

As the pre-trained uni-directional translation model has never seen sentences in the source language on the target side, we first only use it to initialize the subtitling decoder, and use a random initialization for the captioning decoder. To study the effect of initialization, we conduct an ablation study by comparing three settings: initializing only the subtitling decoder, both decoders or the shared decoder (see Table~\ref{tab:res-ft}). Initializing both decoders brings improvements in both directions, with a gain of $1.6$ BLEU for captioning and $0.3$ BLEU for subtitling. Moreover, sharing parameters between decoders further boost the subtitling performance by $0.2$ BLEU. As it seems, the captioning decoder also benefits from a decoder pre-trained in another language.

\begin{table}[!ht]
  \center
  \scalebox{0.85}{
  \begin{tabular}{l|ccc}
  \hline
  Model & EN & FR & Avg\\
  \hline
  \texttt{dual} 1-decoder +FT & 55.3 & 25.3 & 40.3 \\
  \texttt{dual} +FT & 56.9  & 25.6 & 41.3 \\
  \texttt{share} +FT & 56.5 & 25.8 & 41.2 \\
  \hline
  \end{tabular}
  }
  \caption{BLEU scores for multiple initializations.\label{tab:res-ft}}
\end{table}

\subsubsection{Exposure Bias\label{sssec:exposure}}

Due to error accumulations in both decoders, the exposure bias problem seems more severe for dual decoder model than for regular translation models\ \citep{Zhou19synchronous,Zhang20synchronous,Xu21dual}. These authors propose to use \emph{pseudo tri-parallel data with synthetic references} to alleviate this problem. We analyze the influence of this exposure bias issue in our application scenario.

To this end, we compare fine-tuning the \texttt{dual} model with original vs artificial tri-parallel data.
For simplicity, we only report in Table~\ref{tab:res-exposure} the average BLEU scores of captioning and subtitling. Results show that fine-tuning with the original data (w.real) strongly degrades the automatic metrics for the generated text , resulting in performance that are worse than the baseline.

\begin{table}[!ht]
  \center
  \scalebox{0.85}{
  \begin{tabular}{l|ccc}
  \hline
  Model  & Normal & $2$-round & Ref \\
  \hline
  \texttt{dual} +FT w.real   & 39.2 & 40.9 & 45.0 \\
  \texttt{share} +FT w.real & 38.6 & 40.1 & 43.9 \\
  \hline
  \texttt{dual} +FT   & 41.3   & 41.2 & 41.0 \\
  \texttt{share} +FT & 41.2 & 40.9 & 40.5 \\
  \hline
  \end{tabular}
  }
  \caption{Performance of various decoding methods. All BLEU scores are averaged over the two outputs. \textit{$2$-round} (resp.\ \textit{Ref}) refers to decoding with model predictions (resp.\ references) as forced prefix in one direction.\label{tab:res-exposure}}
\end{table}

In another set of experiments, we follow \citet{Xu21dual} and perform asynchronous $2$-round decoding. We first decode the \texttt{dual} models to obtain hypotheses in both languages $\trg_1'$ and $\trg_2'$. During the second decoding round, we use the output English caption $\trg_1'$ as a forced prefix when generating the French subtitles $\trg_2''$. The final English caption $\trg_1''$ is obtained similarly. Note that when generating the $t$-th token in $\trg_2''$, the decoder cross-attention module only attends to the $t$ first tokens of $\trg_1'$, even though the full of $\trg_1'$ is actually known. The $2$-round scores for $\trg_1''$ and $\trg_2''$ are in Table~\ref{tab:res-exposure}, and compared with the optimal situation where we use references instead of model predictions as forced prefix in the second round (in col. `Ref').

Results in Table~\ref{tab:res-exposure} suggest that dual decoder models fine-tuned with original data (w.real) are quite sensible to exposure bias, which can be mitigated with artificial tri-parallel data. Their performance can however be improved by $\sim$$1.5$ BLEU when using $2$-round decoding, thereby almost closing the initial gap with models using synthetic data. The latter approach is overall slightly better and also more stable across decoding configurations.

\section{Conclusion\label{sec:conclusion}}

In this paper, we have explored dual decoding to jointly generate captions and subtitles from ASR transcripts. Experimentally, we found that dual decoding improves translation quality for both captioning and subtitling, while delivering more consistent output pairs. Additionally, we showed that (a) model sharing on the decoder side is viable and effective, at least for related languages; (b) initializing with pre-trained models vastly improves performance; (c) $2$-round decoding allowed us to mitigate the exposure bias problem in our model. In the future, we would like to experiment on more distant language pairs to validate our approach in a more general scenario.

\section{Acknowledgement}

The authors wish to thank Alina Karakanta for providing the ASR transcripts and the evaluation script for the consistency measures. We would also like to thank the anonymous reviewers for their valuable suggestions. This work was granted access to the HPC resources of IDRIS under the allocation 2021-[AD011011580R1] made by GENCI. The first author is partly funded by SYSTRAN and by a grant Transwrite from Région Ile-de-France. This work has also been funded by the BPI-France investment programme "Grands défis du numérique", as part of the ROSETTA-2 project (Subtitling RObot and Adapted Translation).

\bibliography{biblio}

\appendix

\section{Data Processing Details\label{sec:details}}

For the English to French language pair, MuST-Cinema\footnote{License: CC BY-NC-ND 4.0} \citep{Karakanta20must} contains $275$k sentences for training and $1079$ and $544$ lines for development and testing, respectively.
The ASR system used by\ \citet{Karakanta20answer} to produce transcripts was based on the KALDI toolkit \citep{Povey11kaldi}, and had been trained on the clean portion of LibriSpeech \citep{Panayotov15librispeech} ($\sim$$460$h) and a subset of MuST-Cinema ($\sim$$450$h).
In order to emulate a real production scenario, we segment these transcripts as if they were from an ASR system performing segmentation based on prosody.
As this kind of system tends to produce longer sequences compared to typical written text \citep{Cho12segmentation}, we randomly concatenate the English captions into longer sequences, to which we align the ASR transcripts using the conventional edit distance, thus adding a subsegmentation aspect to the translation task.
Edit distance computations are based on a Weighted Finite-State Transducer (WSFT), implemented with Pynini \citep{Gorman16pynini}, which represents editing operations (match, insertion, deletion, replacement) at the character level, with weights depending on the characters and the previous operation context.
After composing the edit WFST with the transcript string and the caption string, the optimal operation sequence is computed using a shortest-distance algorithm \citep{Mohri02semiring}.
The number of sentences to be concatenated is sampled normally, with an average around of $2$.
This process results in $133$k, $499$ and $255$ lines for training, development and testing, respectively.

For pre-training, we use all available WMT14 EN-FR data,\footnote{\url{https://statmt.org/wmt14}} in which we discard sentence pairs with invalid language label as computed by \texttt{fasttext} language identification model\footnote{\url{https://dl.fbaipublicfiles.com/fasttext/supervised-models/lid.176.bin}} \citep{Bojanowski17enriching}. This pre-training data contains $33.9$M sentence pairs. 

\section{Experimental Details\label{sec:expdetails}}

We build our dual decoder model with a hidden size of\ $512$ and a feedforward size of $2048$. We optimize with Adam, set up with a maximum learning rate of $0.0007$ and an inverse square root decay schedule, as well as $4000$ warmup steps.
For fine-tuning, we use Adam with a fixed learning rate of $8\mathrm{e}{-5}$. 
For all models, we share lexical embeddings between the encoder and the input and output decoder matrices.
All models are trained with mixed precision and a batch size of $8192$ tokens on $4$ V100 GPUs. 

The two models in the \texttt{base} setting are trained separately using transcript$\to$caption and transcript$\to$subtitle data. The second model of the \texttt{pipeline} setting is trained using caption$\to$subtitle data. When performing fine-tuning, we first pre-train an EN$\to$FR translation model \texttt{pre-train} using WMT EN-FR data. For \texttt{base}+FT setting, the transcript$\to$subtitle model is fine-tuned from \texttt{pre-train}, while the transcript$\to$caption is the same as \texttt{base} since languages on both source and target sides are English. For \texttt{pipeline}+FT, the caption$\to$subtitle model is fine-tuned from \texttt{pre-train}. For \texttt{dual}+FT, the encoder and the two decoders are fine-tuned from the same \texttt{pre-train} model. The decoder cross-attention layers cannot be fine-tuned and are randomly initialized. Due to computation limits, we are not able to conduct multiple runs for our models. However, all results are obtained by using the parameters averaged over the last $5$~checkpoints.

\section{Consistency Score \label{sec:consistency}}

Consider the following example from \citep{Karakanta21flexibility}:

\vspace{1em}
\noindent
{ \footnotesize
0:00:50,820, 00:00:53,820 \\
To put the assumptions very clearly: \\
\vspace{1em}
Enonçons clairement nos hypothèses : le capitalisme, \\
00:00:53,820, 00:00:57,820 \\
capitalism, after 150 years, has become acceptable, \\
\vspace{1em}
après 150 ans, est devenu acceptable, au même titre \\
00:00:58,820, 00:01:00,820\\
and so has democracy.\\
que la democratie.\\
}\normalsize

As defined by \citet{Karakanta21flexibility}, for the stuctural consistency, both captions (EN) and subtitles (FR) have the same number of $3$ blocks. For lexical consistency, there are 6 tokens of the subtitles which are not aligned to captions in the same block: \textit{``le capitalisme ,'' , ``au même titre''.} 
The $Lex_{C\to S}$ is calculated as the percentage of aligned words normalized by number of words in the caption. 
Therefore, $Lex_{C\to S} = \frac{20}{22} = 90.9\%$; the computation is identical in the other direction, yielding $Lex_{S\to C} = \frac{17}{23} = 73.9\%$, the average lexical consistency of this segment is thus $Lex_{pair} = \frac{Lex_{C\to S} + Lex_{S \to C}}{2} = 82.4\%$.

\begin{table*}[!ht]
  \center
  \scalebox{0.85}{
  \begin{tabular}{l|ccc|ccc|ccc|cc}
  \hline
        & \multicolumn{3}{c|}{TER $\downarrow$} & \multicolumn{3}{c|}{BERTScore-F1 $\uparrow$} & \multicolumn{3}{c|}{BLEU $\uparrow$} & \multicolumn{2}{c}{Consistency $\uparrow$} \\
  Model & EN  & FR  & Avg  & EN  & FR  & Avg  & EN   &  FR   & Avg  & Struct.\ & Lex.\ \\
  \hline
  \texttt{base} & 0.264 & 0.662 & 0.463 & 0.7346 & 0.3961 & 0.5654 & 55.7 & 23.9 & 39.8 & 55.3 & 70.7 \\
  \texttt{base} +FT & 0.264 & 0.654 & 0.459 & 0.7346 & 0.4026 & 0.5686 & 55.7 & 24.9 & 40.3 & 54.5 & 71.4 \\
  \texttt{pipeline} & 0.264 & 0.650 & 0.457 & 0.7346 & 0.3912 & 0.5629 & 55.7 & 23.6 & 39.7 & 95.7 & 96.0 \\
  \texttt{pipeline} +FT & 0.264 & 0.652 & 0.458 & 0.7346 & 0.3924 & 0.5635 & 55.7 & 24.2 & 40.0 & 98.4 & 98.3 \\
  \texttt{dual} +FT & \textbf{0.256} & \textbf{0.640} & \textbf{0.448} & 0.7378 & \textbf{0.4074} & 0.5726 & \textbf{56.9} & 25.6 & \textbf{41.3} & 65.1 & 79.1 \\
  \texttt{share} +FT & 0.259 & \textbf{0.640} & 0.450 & \textbf{0.7396} & 0.4066 & \textbf{0.5731} & 56.5 & \textbf{25.8} & 41.2 & \textbf{66.7} & \textbf{80.0} \\
  \hline
  \end{tabular}
  }
  \caption{TER, BERTScore and BLEU scores for captions (EN) and subtitles (FR), with measures of structural and lexical consistency between the two hypotheses. The \texttt{base} and \texttt{pipeline} settings are trained from scratch with original data. \texttt{share} refers to tying all decoder parameters. Signature of BERTScore (EN): microsoft/deberta-xlarge-mnli\_L40\_no-idf\_version=0.3.11(hug\_trans=4.10.3)-rescaled\_fast-tokenizer. Signature of BERTScore (FR): bert-base-multilingual-cased\_L9\_no-idf\_version=0.3.11(hug\_trans=4.10.3)-rescaled\_fast-tokenizer.\label{tab:res-bert}}
\end{table*}

\begin{table*}[!ht]
  \center
  \scalebox{0.85}{
  \begin{tabularx}{1.1\linewidth}{l|X}
  \hline
  Source & take time to write down your values your objectives and your key results do it today \\
  \hline
  EN \texttt{pipeline} +FT & Take time to write down [eol] your values, your objectives, [eob] and your key results do it today. [eob] \\
  EN \texttt{share} +FT & Take time to write down your values, \textbf{[eol]} your objectives, [eob] and your key results do it today. [eob] \\
  EN ref & Take time to write down your values, [eob] your objectives and your key results. [eob] Do it today. [eob] \\
  \hline
  FR \texttt{pipeline} +FT & Prenez le temps d'écrire vos valeurs, [eol] vos objectifs, [eob] et vos principaux résultats [eol] le font aujourd'hui. [eob] \\
  FR \texttt{share} +FT & Prenez le temps d'écrire vos valeurs, \textbf{[eob]} vos objectifs et \textbf{vos résultats clés. [eob] Faites-le} aujourd'hui. [eob] \\
  FR ref & Prenez le temps d'écrire vos valeurs, [eob] vos objectifs et vos résultats clés. [eob] Faites-le aujourd'hui. [eob] \\
  \hline
  \hline
  Source & and as it turns out what are you willing to give up is exactly the right question to ask \\
  \hline
  EN \texttt{pipeline} +FT & And as it turns out, what are you willing [eol] to give up is exactly [eob] the right question to ask? [eob] \\
  EN \texttt{share} +FT & And as it turns out, what are you willing [eol] to give up \textbf{[eob]} is exactly the right question to ask? [eob] \\
  EN ref & And as it turns out, [eob] "What are you willing to give up?" [eob] is exactly the right question to ask. [eob] \\
  \hline
  FR \texttt{pipeline} +FT & Et il s'avère que ce que vous voulez abandonner [eol] est exactement [eob] la bonne question à poser ? [eob] \\
  FR \texttt{share} +FT & Et il s'avère que ce que vous voulez abandonner \textbf{[eob]} est exactement la bonne question à poser\textbf{.} [eob] \\
  FR ref & Et il s'avère que [eob] « Qu'êtes-vous prêts à abandonner ? » [eob] est exactement la question à poser. [eob] \\
  \hline
  \end{tabularx}
  }
  \caption{Examples of dual decoding improving both captioning and subtitling. Major improvements are marked in bold.\label{tab:example}}
\end{table*}

When computing the \textit{lexical consistency} between captions and subtitles, we use the WMT14 EN-FR data to train an alignment model using \texttt{fast\_align}\footnote{\url{https://github.com/clab/fast\_align}} \citep{Dyer13simple} in both directions and use it to predict word alignments for model outputs.

\section{Additional Metric}\label{sec:metric}

Table~\ref{tab:res-bert} reports TER and BERTScores\footnote{\url{https://github.com/Tiiiger/bert_score}} \citep{Zhang20BERTScore}. Note that for BERTScores, we remove segmentation tokens ([eob] and [eol]) from hypotheses and references, as special tokens are out-of-vocabulary for pre-trained BERT models.

\section{Examples\label{sec:example}}

Some examples of dual decoding improving the quality of both captioning and subtitling compared to the pipeline system are in Table~\ref{tab:example}.

\end{document}